\documentclass[10pt,twocolumn,letterpaper]{article}

\usepackage[pagenumbers]{wacv} 

\usepackage{graphicx}
\usepackage{amsmath}
\usepackage{amssymb}
\usepackage{booktabs}
\usepackage{array}
\usepackage{multirow}
\usepackage{rotating}
\usepackage{makecell}
\usepackage{textcomp}
\usepackage{tikz}
\usepackage{enumitem}

\usetikzlibrary{spy}
\usetikzlibrary{calc}

\usepackage{xcolor}

\definecolor{MaterialLoss}{HTML}{1CE6FF}
\definecolor{Peel}{HTML}{FF34FF}
\definecolor{Dust}{HTML}{FF4A46}
\definecolor{Scratch}{HTML}{008941}
\definecolor{Hair}{HTML}{006FA6}
\definecolor{Dirt}{HTML}{A30059}
\definecolor{Fold}{HTML}{FFA500}
\definecolor{Writing}{HTML}{7A4900}
\definecolor{Cracks}{HTML}{0000A6}
\definecolor{Staining}{HTML}{63FFAC}
\definecolor{Stamp}{HTML}{004D43}
\definecolor{Sticker}{HTML}{8FB0FF}
\definecolor{Puncture}{HTML}{997D87}
\definecolor{Background}{HTML}{5A0007}
\definecolor{BurnMarks}{HTML}{809693}
\definecolor{Lightleak}{HTML}{F6FF1B}
\definecolor{GHOST}{HTML}{FFFFFF}

\renewcommand{\paragraph}[1]{\par\noindent\textbf{#1}~}

\usepackage[pagebackref,breaklinks,colorlinks]{hyperref}

\usepackage[capitalize]{cleveref}
\crefname{section}{Sec.}{Secs.}
\Crefname{section}{Section}{Sections}
\Crefname{table}{Table}{Tables}
\crefname{table}{Tab.}{Tabs.}

\usepackage[accsupp]{axessibility}

\def\wacvPaperID{1470} 
\def\confName{WACV}
\def\confYear{2025}

\begin{document}

\title{ARTeFACT: Benchmarking Segmentation Models on Diverse Analogue Media Damage}

\author{Daniela Ivanova\\
University of Glasgow,\\
UK\\
\and
Marco Aversa\\
Dotphoton,\\
Switzerland\\
\and
Paul Henderson\\
University of Glasgow,\\
UK\\
\and
John Williamson\\
University of Glasgow,\\
UK\\
}
\maketitle

\begin{figure*}[!htbp]
    \centering
    \includegraphics{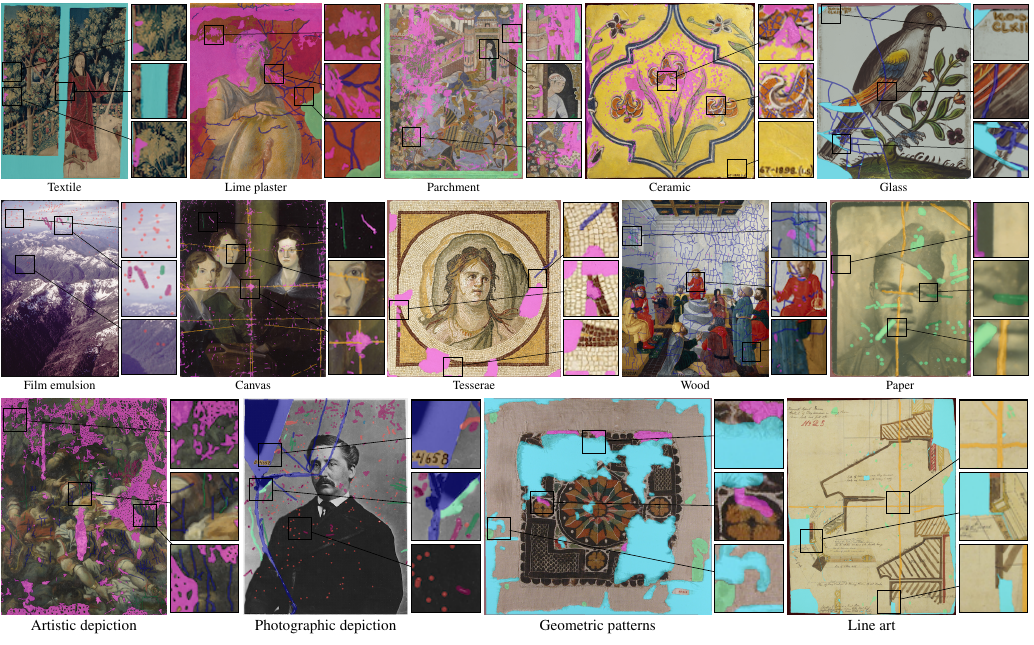}
        \vspace{-8pt}
\caption{Examples from our dataset of damaged artwork, categorised by Material (rows 1 and 2) and Content (row 3). Annotation colours correspond to different types of damage. Note the diversity of media and content, and pixel-accurate damage masks.}
\label{fig:categories}
\end{figure*}

\begin{abstract}
Accurately detecting and classifying damage in analogue media such as paintings, photographs, textiles, mosaics, and frescoes is essential for cultural heritage preservation. While machine learning models excel in correcting degradation if the damage operator is known a priori, we show that they fail to robustly predict \emph{where} the damage is even after supervised training; thus, reliable damage detection remains a challenge. Motivated by this, we introduce ARTeFACT, a dataset for damage detection in diverse types analogue media, with over 11,000 annotations covering 15 kinds of damage across various subjects, media, and historical provenance. Furthermore, we contribute human-verified text prompts describing the semantic contents of the images, and derive additional textual descriptions of the annotated damage. We evaluate CNN, Transformer, diffusion-based segmentation models, and foundation vision models in zero-shot, supervised, unsupervised and text-guided settings, revealing their limitations in generalising across media types. Our dataset is available at \href{https://daniela997.github.io/ARTeFACT/}{https://daniela997.github.io/ARTeFACT/} as the first-of-its-kind benchmark for analogue media damage detection and restoration.
\end{abstract}

\section{Introduction}
\label{sec:intro}

Artworks on analogue media are at risk of deterioration over time due to environmental factors, human intervention, or simple aging. Accurately distinguishing and analysing such damage is a crucial step in the preservation process, with many practical applications---including improved archiving and curation, provenance research, and downstream restoration via digital tools. Manual identification of damage is labour-intensive, requiring specialized software, significant financial investment, and extensive time commitment from skilled specialists~\cite{chambah2005, calatroni2018unveiling, ivanova23analogue}. These constraints limit the scale and frequency of restoration efforts.

Whether machine learning can successfully be used to automate the damage detection has not yet been established. This can be attributed to the constraints posed by data availability: damage in analogue media is rarely detailed in metadata, making data collection challenging. Selecting and annotating examples which cover the varied manifestations of damage across different materials and contents in analogue media is difficult but essential for assessing a model's ability to generalise at such a complex task~\cite{chambah2019}. Previous work has only targeted one type of analogue media at a time and sourced all of their data from the same place~\cite{calatroni2018unveiling, liu2022digital, yuan2023automatic, merizzi2024deep}. Consequently, their results do not reflect the evaluated approaches' performance on unseen data, and whether the models are able to truly learn what damage \emph{is}. Effective evaluation requires a diverse dataset and a protocol that accounts for the manifold of analogue media types and their differences.

We propose the ARTeFACT dataset, which is designed to enable a comprehensive evaluation of current and future approaches.
Our contributions are as follows:
\begin{itemize}[itemsep=0pt, topsep=0.3pt, parsep=0.2pt]
    \item We produce an extensive dataset for the real-life task of damage detection in analogue media. For each image, we provide a pixel-accurate damage mask, with over 11,000 annotations in total, across 418 high-resolution images spanning various cultures and historical periods. This dataset is the first of its kind.
    \item We propose and justify a comprehensive taxonomy of analogue damage, categorizing deterioration into 15 distinct classes. We classify the images into 10 material categories to represent how damage manifests across different media based on material properties. Additionally, we use 4 content categories to capture differences in content complexity and structure.
    \item We perform a thorough evaluation of state-of-the-art segmentation models in various settings - including a vision foundation model trained specifically for segmentation - and demonstrate that all models fall short of generalising across different analogue media domains and damage types. 
    \item We further benchmark two state-of-the-art diffusion-based segmentation methods. Our results suggest that current text-to-image models have insufficient specificity in conditioning for damage, resulting in the segmentation of irrelevant areas. 
\end{itemize}

\section{Related work}
\label{sec:related-work}

\paragraph{Damage in analogue media.}
Advances in digital technology have revolutionized access to cultural artifacts by enabling digitization. Digitization creates durable replicas, but also captures \textit{damage} that was present on the original media.
The field of cultural heritage preservation has faced ongoing challenges in defining what \emph{is} and \emph{isn't} damage, and hence - what should and should not be restored~\cite{stoner2005changing}, complicating conservators' decisions on restoration scope and methods\cite{oddy1999reversibility}. By detecting and restoring damage digitally, we can address and mitigate the risk of causing further harm to the original analogue medium; this aligns with the crucial conservation principle of reversibility \cite{beck1999reversibility, oddy1999reversibility}. However, this also necessitates skilled use of specialised software by restoration professionals \cite{chambah2005, chambah2006, chambah2019}. Damage in analogue media, which includes a range of materials such as film, paintings, mosaics, murals, drawings, textile and others, can manifest in diverse forms. Chambah \cite{chambah2005} categorizes such damage broadly into two groups: chemical degradations and mechanical degradations. While this classification found in Chambah's work pertains to film emulsion \cite{chambah2006, chambah2019}, similar ontologies are commonly applied in the field of mathematical modelling of painting damage \cite{de1999mathematical, pauchard2020craquelures, calatroni2018unveiling, dang2021mathematical}. 

\paragraph{Image restoration and segmentation.}
The task of image restoration typically involves reversing a degradation process $x$ applied to the undamaged image $I$ to produce the damaged image $I^\prime$. Common restoration tasks include denoising \cite{pmlr-v80-lehtinen18a, zamir2021multi, Zamir_2022_CVPR, liang2021swinir, chen2022simple}, superresolution \cite{johnson2016perceptual, wang2018esrgan, liang2021swinir}, and colourisation \cite{zhang2016colorful, isola2017image, kumar2021colorization, saharia2022palette}. Training data is often generated by applying transformations to undamaged images, like creating grayscale versions for colourisation. These tasks are primarily digital and easy to simulate, unlike analogue media damage, which stems from the physical properties of the medium and is only digitised after photography or scanning. The irregular nature of such damage—material loss, cracks, scratches—presents significant challenges for digital simulation \cite{chambah2019, wan2020bringing, mironicua2020generative, visapp22, ivanova23analogue, calatroni2018unveiling, sankar2023transforming, cornelis2013crack, dang2021mathematical}.

In most restoration tasks, such as superresolution and colourisation, the degradation operator is global; when damage is localized (e.g., inpainting), the damaged areas are typically assumed to be known. Inpainting has been addressed using VAEs \cite{ham18, peng2021generating}, GANs \cite{yu2018generative}, and diffusion models \cite{saharia2022palette, rombach2021highresolution, lugmayr2022repaint, yang2023pgdiff, wang2022zero}. Stable Diffusion \cite{rombach2021highresolution} has demonstrated state-of-the-art inpainting capabilities and is now the go-to tool for conditional image inpainting. However, these models rely on masks indicating the damaged areas. This suggests that research should shift focus from inpainting to \emph{detecting} damage, the most challenging and understudied part of the machine learning pipeline.

Blind inpainting, where the mask indicating the damaged area is not provided, typically requires a segmentation module to detect these areas, using models like autoencoders \cite{wang20eccv,cai17tvc,hertz2019blind} or U-Nets \cite{mironicua2020generative, ivanova23analogue}. State-of-the-art architectures like UPerNet~\cite{xiao2018unified} and SegFormer~\cite{xie2021segformer} could be trained for damage segmentation if annotations are available. Self-supervised models such as DINO~\cite{caron2021emerging}, DINOv2~\cite{oquab2023dinov2}, SAM~\cite{kirillov2023segment}, and CLIP~\cite{radford2021learning} have also shown promise in learning semantically rich representations useful for segmentation tasks. However, their ability to generalise to the complex task of damage segmentation, with its variability in shape and distribution, has remained an open question, which we address in this work.

Recent text-to-image diffusion models like Stable Diffusion~\cite{rombach2021highresolution} have inspired methods such as DiffEdit~\cite{couairon2022diffedit}, which uses pre-trained diffusion models to detect damaged areas based on text prompts, followed by inpainting. DiffSeg~\cite{tian2024diffuse}, the state-of-the-art for unsupervised segmentation, leverages diffusion models for semantic segmentation without additional supervision.

\paragraph{Datasets.}
Several art datasets have been developed for classification~\cite{Crowley14, saleh2015large} and semantic segmentation~\cite{cohen2022semantic} in cultural heritage. More recent datasets pair artwork images with textual descriptions~\cite{conde2021clip, Bleidt_2024_WACV}, and even leverage text conditioning to synthesise artwork variants via diffusion models~\cite{cioni2023diffusion}. However, there is a notable scarcity of datasets of damaged analogue media, and the few that do exist tend to focus exclusively on one type of media: paintings~\cite{cornelis2013crack, sankar2023transforming}, illuminated manuscript miniatures~\cite{calatroni2018unveiling}, film emulsion scans~\cite{ivanova23analogue}, frescoes~\cite{merizzi2024deep}. 

\section{Damaged Analogue Media Dataset}\label{dataset}

\begin{figure}[t] 
\begin{center}
    \includegraphics{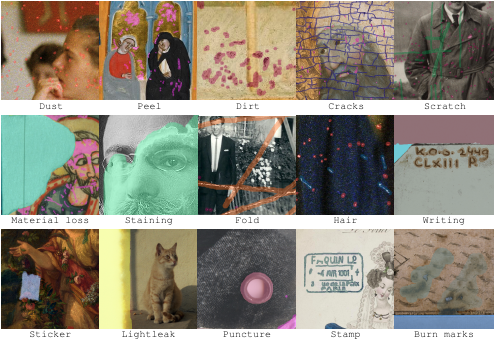}
\end{center}
\vspace{-1.5em}
\caption{Damage types found in our dataset, demonstrating the variety of shape, scale and severity of damage.}
\vspace{-1em}
\label{fig:damage-examples}
\end{figure}

We present a new dataset for damage detection in analogue media. Our dataset consists of 418 high-resolution images of various kinds of analogue media, including examples of: manuscript miniatures, photographs, maps, ephemera (such as posters and leaflets), mosaics, drawings and sketches, paintings, frescoes, carpets, tiles, lithographs, book covers, technical drawings and blueprints, wallpapers, letters, tapestries and stained glass (indicated as a \texttt{Type} attribute in the dataset's metadata). We provide over 11,000 pixel-level annotations covering 15 types of damage, and categorise them based on material and content. In addition to these annotations, we also make available natural language annotations, disentangled into content and damage descriptions.
Images were manually selected from digitized collections of museums and galleries, with additional sources from WikiArt and Flickr groups, all collected under a CC license at the highest possible resolution.

\subsection{Annotation methodology}
All images were annotated with 15 physical damage types by one expert annotator, followed by two rounds of manual review to ensure quality. A second expert conducted a final review, resulting in over 11,000 annotated instances. Our expert annotators specialise in digitally restoring analogue media and have experience with all media types in the dataset. Each image is also classified by material and content based on its metadata, yielding 10 material classes and 4 content classes. For a detailed look at our images and image-level annotations through the prism of CLIP, refer to Section~ in the Supplementary.

\begin{figure}[t]
    \centering
    \begin{subfigure}{0.92\linewidth}
        \centering
        \includegraphics[width=\linewidth,trim={0em 0em 0em 0em},clip]{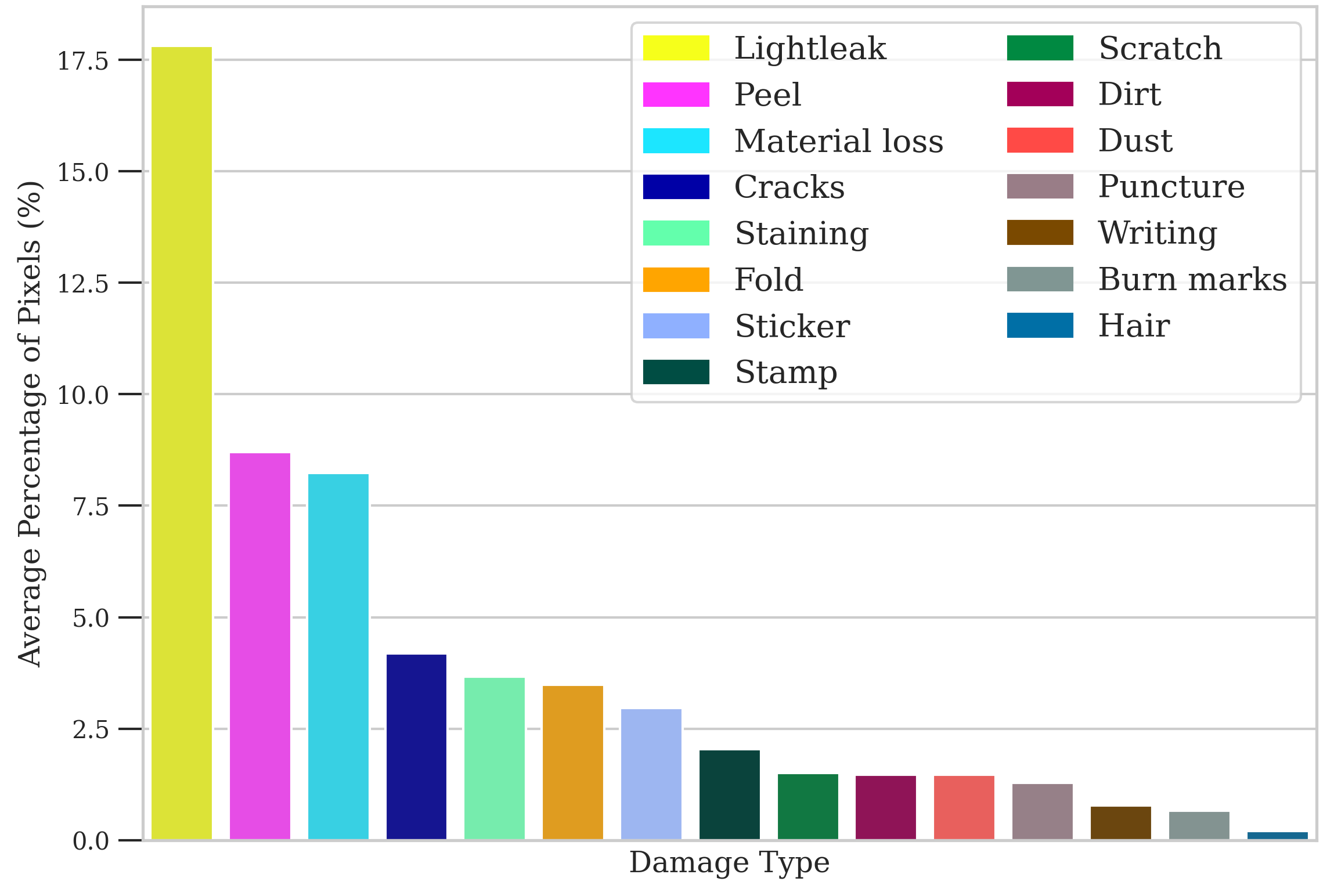}
        \caption{Severity of Damage Types in Affected Images}
        \label{fig:average-damaged-pixels}
    \end{subfigure}
    \hfill
    \begin{subfigure}{0.92\linewidth}
        \centering
        \includegraphics[width=\linewidth,trim={0em 0em 0em 0em},clip]{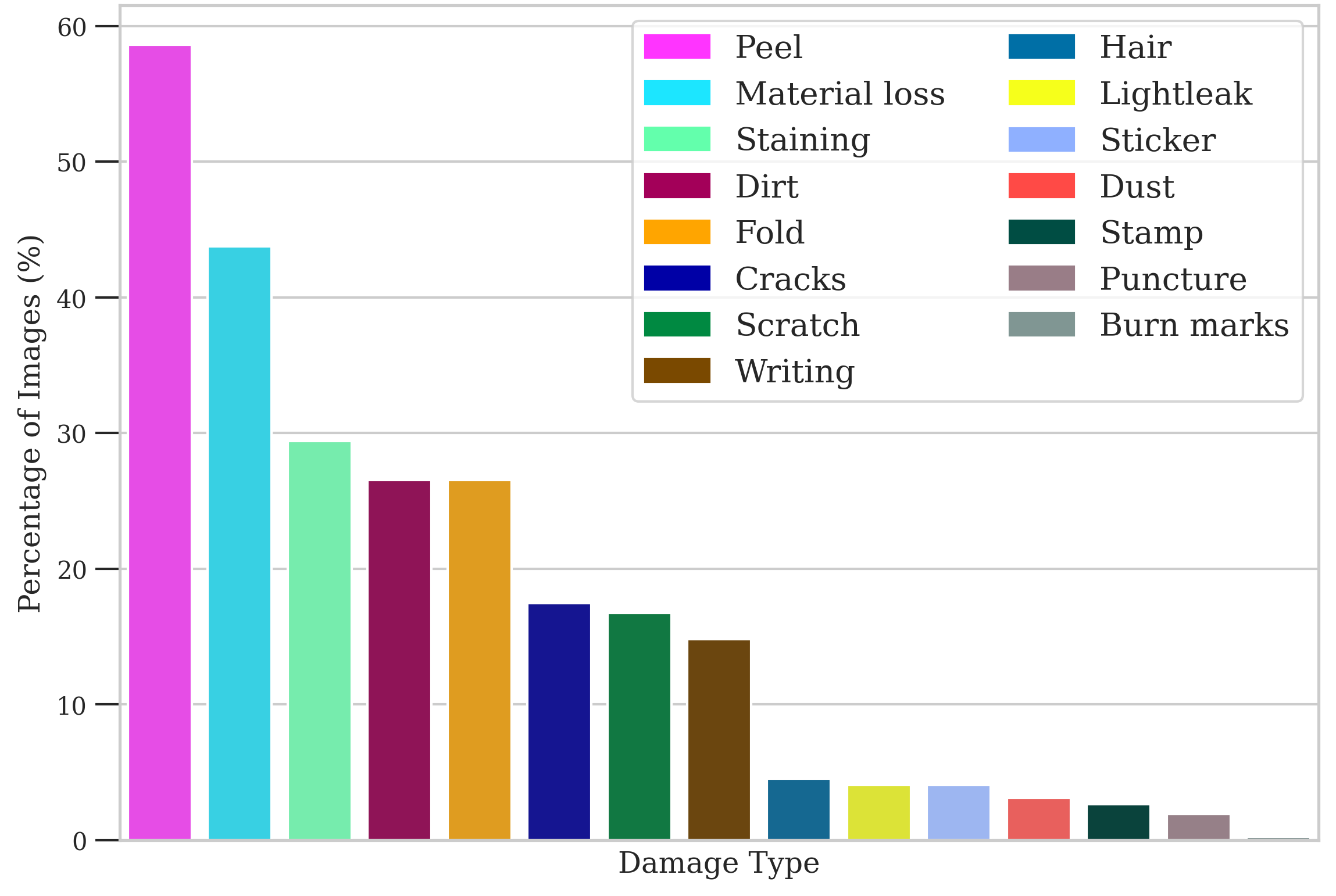}
        \caption{Prevalence of Damage Types in the Dataset}
        \label{fig:damage-distribution}
    \end{subfigure}
    \caption{Overview of the prevalence and severity of different damage types in the Dataset.}
    \vspace{-17pt}
    \label{fig:damage-types}
\end{figure}

\subsection{Damage Types}
We define damage types by their properties (scale, opacity), causes (mechanical, chemical), and effects (occluding information, absence of information, structural deformity). Each pixel-level annotation is assigned a damage type from the proposed taxonomy, with \texttt{Clean} denoting parts of the artwork which are undamaged, and remaining pixels classified as \texttt{Background}, covering the framing surface of digitisation, if visible (usually in the cases of non-rectangular artwork). This taxonomy was developed in consultation with multiple heritage preservation experts. 

The types of damage include \texttt{Material loss}, which refers to missing or eroded sections of the artwork, and \texttt{Peel}, indicating areas where layers (e.g.~of pigment) have separated or are flaking off. \texttt{Dust}, \texttt{Hair}, and \texttt{Dirt} denote different kinds of surface contaminants. \texttt{Scratch}, \texttt{Puncture}, \texttt{Fold} and \texttt{Cracks} capture mechanical damage and deformities, potentially caused by mishandling or accidents, or due to age. \texttt{Stamp} and \texttt{Sticker} highlight intentional or accidental markings that could be seen as defacements. \texttt{Staining} and \texttt{Burn marks} are indicative of chemical or thermal damage. \texttt{Lightleak} is a specific type of damage often found in photographic materials, caused by unintended exposure to light. 

Damage types vary widely in terms of shape, size and spatial distribution; Figure~\ref{fig:damage-types} shows the distribution and prevalence of each type of damage within the dataset. Figure~\ref{fig:average-damaged-pixels} considers the extent of damage as a percentage of pixels covered. In Figure~\ref{fig:damage-distribution}, each distinct type of damage present in an image is instead counted as a single instance, regardless of its extent or interconnectedness. We can see that Peel and Material Loss are the most common types of damage found in the dataset, whereas Lightleaks are less common. At the same time, Lightleaks are very severe (i.e.~cover many pixels) when they occur.  Dirt, Fold, and Staining are common but vary in severity. Further insight can be gained by categorising the images in the dataset by material and content.

\subsection{Image-level Annotations}

In addition to the pixel-level damage annotations, we also provide three types of image-level annotations.

\paragraph{Material categorisation.}
How damage manifests itself is correlated to the materials used in the imaging process; each material type has unique properties and susceptibilities to different forms of damage. We propose the following categories, based on information sourced from the images' metadata during the collection process: \texttt{Parchment}, \texttt{Film emulsion}, \texttt{Glass}, \texttt{Paper}, \texttt{Tesserae}, \texttt{Canvas}, \texttt{Lime plaster}, \texttt{Textile}, \texttt{Ceramic}, \texttt{Wood}; examples for each category are shown in Figure~\ref{fig:categories}, rows 1 and 2.

\paragraph{Content categorisation.}
Besides classifying based on material, we also provide content categorisation to capture the differences of overall image attributes across analogue media types, which can vary in complexity and style---for instance, photographs capture natural scenes, while paintings are stylised depictions, tile and carpet designs usually consist of repetitive geometric or abstract patterns. The following categories are proposed based on the observed differences in types content in each type of imagery: \texttt{Artistic depiction}, \texttt{Photographic depiction}, \texttt{Line art}, \texttt{Geometric patterns}; examples for each category in Figure~\ref{fig:categories}, row 3.

\paragraph{Textual descriptions.}
We provide textual descriptions for each image, detailing both content and damage types. The expert descriptions are derived by correcting draft captions produced by LLaVA~\cite{liu2023llava}. Additional damage descriptions were compiled from our pixel-level annotations. For detailed examples and discussion on how much the expert captions improve those of LLaVA refer to Section~2. in the Supplementary.

\begin{figure*}[t]
\begin{center}
\includegraphics{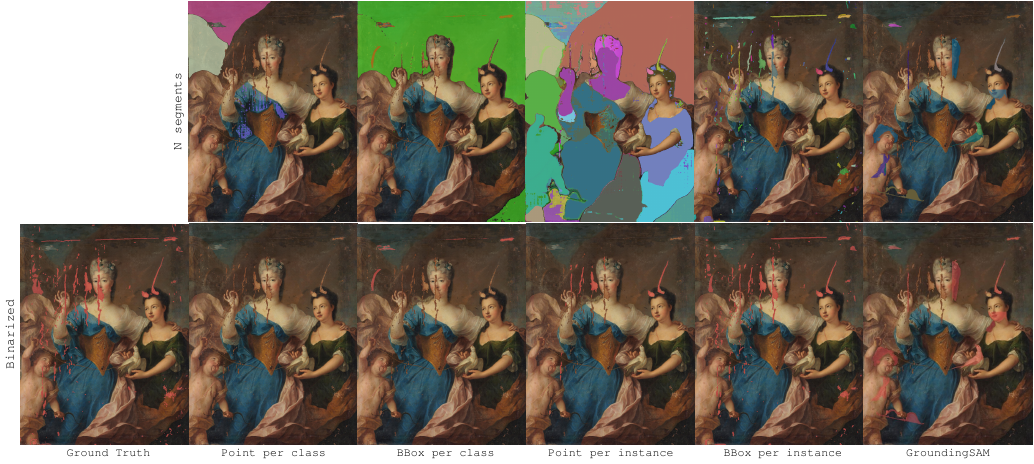}
\end{center}
\vspace{-10pt}
\caption{Qualitative results for SAM at zero-shot damage segmentation. Top row shows initial segments from N prompts as predicted by SAM (unique colour per prompt), bottom row shows the segments after being assigned binary class (Clean or Damaged) via an oracle.}
\label{fig:sam-zero-shot}
\end{figure*}

\textbf{\begin{table*}[t]
\tiny
\caption{Results from benchmarking SAM at zero-shot damage segmentation across various prompt modes. N corresponds to average number of prompts (i.e. points, bounding boxes, text) required as input. Point grid was evaluated with 64 points, but those do not need to be provided a priori.}
\label{tab:sam}
\vspace{-8pt}
\resizebox{\linewidth}{!}{%
\begin{tabular}{@{}lcccccccccccccccccccccccc@{}lll}
\toprule
\multirow{2}{*}{\textbf{Test Class}} & \multicolumn{4}{c}{\textbf{Point per class}}                     & \multicolumn{4}{c}{\textbf{BBox per class}}     & \multicolumn{4}{c}{\textbf{Point per instance}}            & \multicolumn{4}{c}{\textbf{BBox per instance}} &                   \multicolumn{4}{c}{\textbf{Point Grid}} &                   \multicolumn{4}{c}{\textbf{Grounding SAM}}\\ 
\cmidrule(l){2-5} \cmidrule(l){6-9} \cmidrule(l){10-13} \cmidrule(l){14-17} \cmidrule(l){18-21} \cmidrule(l){22-25} 
                            & \textbf{Acc} & \textbf{F1} & \textbf{mIoU} & \textbf{N} & \textbf{Acc} & \textbf{F1} & \textbf{mIoU} & \textbf{N} & \textbf{Acc} & \textbf{F1} & \textbf{mIoU} & \textbf{N} & \textbf{Acc} & \textbf{F1} & \textbf{mIoU} & \textbf{N} & \textbf{Acc} & \textbf{F1} & \textbf{mIoU} & \textbf{N} & \textbf{Acc} & \textbf{F1} & \textbf{mIoU} & \textbf{N} \\ 
                            \midrule
\textit{Wood}          &  0.85   &  0.15   &  0.10 & 2 &  0.86    &  0.15   &  0.11   & 2  &  0.88   &  0.41  &   0.31 & 206  &  0.90   &    \textbf{0.60}   &   0.49 & 206  & 0.87    &  0.41  &  0.32 & - &  0.83  &   0.05  & 0.03 & 2\\
\textit{Ceramic}            &    0.85 &  0.39  &   0.27 & 4  &  0.79   &  0.09  &   0.06  & 4 &  0.85   &  0.50  &   0.35 & 185  &  0.89   &  \textbf{0.57}      &   0.44  & 185  & 0.86    &  0.33  & 0.25 & - &  0.67 & 0.03 & 0.01 & 4\\
\textit{Textile}            &  0.89   &  0.39  &   0.32  & 2 &  0.85   &  0.16  &   0.15 & 2  &   0.93  &  0.68  &   0.56  & 76 &  0.91   &    \textbf{0.68}    &   0.57  & 76 &  0.87   &  0.37   &  0.28  & - &   0.87 & 0.03 & 0.01 & 2\\
\textit{Lime Plaster}       &   0.87  &  0.41  &   0.32  & 3 &  0.85   &  0.31  &   0.25  & 3 &   0.90  &  0.57  &   0.46  & 90 &  0.90   &    \textbf{0.67}    &  0.56 & 90 & 0.90     &  0.60  &  0.49 & -  &  0.60 & 0.02 & 0.01 & 3\\
\textit{Canvas}             &  0.88   &  0.19  &   0.15   & 2 &  0.88   & 0.14   &   0.10   & 2 &  0.90   &  0.33  &   0.26  & 95  &  0.90   &    \textbf{0.57}    &  0.44 & 95 & 0.89     &  0.32  & 0.24 & -  &  0.93 & 0.21 & 0.12 & 2\\
\textit{Tesserae}           &  0.89   &  0.43  &  0.36  & 2 &   0.86  & 0.26   &  0.23   & 2 &   0.90  &  0.59  &   0.49 & 43 &  0.93   &   \textbf{0.72}     &  0.62  & 43 &  0.91   & 0.60   &  0.49 & -  &  0.77 & 0.03 & 0.01 & 2\\
\textit{Paper}              &   0.92  &  0.36  &  0.28  & 3  &  0.91   & 0.23   &   0.19 & 3  &  0.93   &  0.44  &   0.34 & 37  &  0.95   &   \textbf{0.60}     &  0.49 & 37 & 0.93    &  0.45  &   0.35 & - &  0.86 & 0.05 & 0.02 & 3 \\
\textit{Glass}              &   0.91  &  0.35  &   0.26  & 5 &  0.90   &  0.23  &  0.18  & 5  &  0.92   &  0.49  &   0.38  & 335  &  0.93   &   \textbf{0.63}     &  0.50 & 335  &0.93     &  0.47  & 0.37 & -  &  0.89 & 0.10 & 0.05 & 5\\
\textit{Film emulsion}      &   0.94  &  0.38  &   0.35   & 2 &   0.95  &  0.62  &   0.54 & 2  &  0.95   & 0.52   &  0.45  & 261  &  0.97   &  \textbf{0.79}  &  0.69  & 261 & 0.93   & 0.55      &  0.47 & -  &   0.94 & 0.08 & 0.04 & 2\\
\textit{Parchment}          &   0.92  &  0.18  &   0.13   &  3 &  0.92  & 0.13   &  0.10  & 3 &  0.93   &  0.31  &  0.24 & 170   &  0.94   &  \textbf{0.58}    &   0.45  & 170 &  0.93 &  0.31   &  0.23  & -  &    0.88 & 0.06 & 0.03 & 3\\
\midrule
\textit{Geom Patterns}      &   0.90  &  0.43  &   0.35   & 3 &   0.86  & 0.12   &   0.10  & 3 &  0.91   &  0.58  &   0.47  & 100 &  0.92   &  \textbf{0.65}   &  0.54 & 100  & 0.89   &  0.40  &    0.31  & -   &  0.82 & 0.04 & 0.02 & 3\\
\textit{Line Art}           &   0.93  &  0.33  &   0.23   & 3 &  0.93  & 0.20   &   0.16  & 3 &   0.95  &  0.46  &  0.36  & 38  &  0.95   &    \textbf{0.56}    &  0.45  & 38  & 0.95    &  0.44  &  0.33 & - & 0.92 & 0.05 & 0.03 & 3 \\
\textit{Photo Depiction} &  0.92  &  0.36   &  0.27  &  3 &  0.91   &   0.30  & 0.24   &   3 &   0.93&     0.45&   0.35 & 163 & 0.94  &  \textbf{0.63}     &  0.51 & 163  & 0.93   &  0.45  &    0.35 & -  &   0.91 & 0.11 & 0.06 & 3\\
\textit{Art Depiction}       &  0.90   &  0.31   &  0.23  & 3 &   0.89  &   0.21  &  0.17  &  3 &   0.91 &    0.44 &    0.35 & 93 & 0.93   &  \textbf{0.63}    &  0.51 & 93 & 0.91   &  0.45   &   0.36 & -  &  0.80 & 0.04 & 0.02 & 3\\ \midrule
\textit{Overall}   &   0.91  &  0.33  &   0.25  & 3 &  0.90   &  0.23  & 0.19   & 3 &  0.92 &     0.46 &    0.36 &  112 & 0.93    &   \textbf{0.63} & 0.52 & 112 &  0.92 &  0.45  & 0.32 & - &  0.85  &  0.05   &   0.02 & 3 \\ \bottomrule
\end{tabular}%
}
\end{table*}
}

\begin{table*}[t]
\tiny
\centering
\caption{Results across both Material and Content splits, for all baselines, trained on the task of binary damage segmentation. Best F1 scores for each test class in bold.}
\label{tab:baselines-binary-semantic-segmentation}
\vspace{-8pt}
\resizebox{\linewidth}{!}{%
\begin{tabular}{@{}lcccccccccccccccccc@{}}
\toprule
\multirow{2}{*}{Test Class} & \multicolumn{3}{c}{\textbf{Segformer}}                     & \multicolumn{3}{c}{\textbf{\makecell{UPerNet  + Swin}}}     & \multicolumn{3}{c}{\textbf{\makecell{UPerNet  + ConvNeXt}}}            & \multicolumn{3}{c}{\textbf{\makecell{DINOv2  + MLP}}} &                   \multicolumn{3}{c}{\textbf{\makecell{SAM ViT-H + MLP}}}\\ 
\cmidrule(l){2-4} \cmidrule(l){5-7} \cmidrule(l){8-10} \cmidrule(l){11-13} \cmidrule(l){14-16}
                            & \textbf{Acc} & \textbf{F1} & \textbf{mIoU} & \textbf{Acc} & \textbf{F1} & \textbf{mIoU} & \textbf{Acc} & \textbf{F1} & \textbf{mIoU} & \textbf{Acc} & \textbf{F1} & \textbf{mIoU}  & \textbf{Acc} & \textbf{F1} &\textbf{mIoU}  \\ \midrule
\textit{Wood}               & 0.85& 0.48& 0.32& 0.84& 0.36& 0.22& 0.85& \textbf{0.59}& 0.23& 0.85& 0.47& 0.31 &0.81 &0.40 &0.25\\
\textit{Ceramic}            & 0.87& 0.46   & 0.46& 0.82& 0.43& 0.27& 0.83& 0.46                  & 0.30& 0.82& \textbf{0.61}                  & 0.44             & 0.80& 0.37& 0.23\\
\textit{Textile}            & 0.83& 0.49& 0.33& 0.86& \textbf{0.60}& 0.43& 0.88& \textbf{0.60}                  & 0.43& 0.89& 0.59                  & 0.42             & 0.84& 0.48& 0.30\\
\textit{Lime Plaster}       & 0.85& \textbf{0.66}& 0.49& 0.84& 0.58& 0.40& 0.83& 0.64                  & 0.48& 0.82& 0.61                  & 0.44             & 0.79& 0.52& 0.35\\
\textit{Canvas}             & 0.93& \textbf{0.70}& 0.53& 0.92& 0.69& 0.52& 0.92& 0.67                  & 0.50& 0.90& 0.62                  & 0.45             & 0.82 & 0.50 & 0.33\\
\textit{Tesserae}           & 0.85& \textbf{0.54}& 0.37& 0.81& 0.47& 0.31& 0.84& 0.47                  & 0.31& 0.85& 0.52                  & 0.36             & 0.79 & 0.47 & 0.30\\
\textit{Paper}              & 0.92& 0.56& 0.39& 0.90& 0.53& 0.35& 0.92& \textbf{0.59}                  & 0.42& 0.89& 0.50                  & 0.34             &0.84 & 0.44 & 0.28\\
\textit{Glass}              & 0.89& 0.30& 0.17& 0.89& 0.37& 0.22& 0.90& 0.44                  & 0.28& 0.88& \textbf{0.46}                  & 0.30    & 0.86&0.36 &0.22\\
\textit{Film emulsion}      & 0.85& 0.34& 0.20& 0.85& 0.50& 0.33& 0.93& \textbf{0.71}                 & 0.56& 0.86& 0.46                  & 0.30             & 0.89&0.63 &0.46\\
\textit{Parchment}          & 0.89& \textbf{0.44}& 0.28& 0.90& \textbf{0.44}& 0.28& 0.89& 0.35                  & 0.21& 0.88& 0.35                  & 0.21             & 0.84&0.32 &0.19\\
\midrule
\textit{Geom Patterns}      & 0.90& \textbf{0.64}& 0.48& 0.89& 0.61& 0.44& 0.84& 0.57                  & 0.39& 0.89& 0.62                  & 0.45             & 0.86& 0.52& 0.35\\
\textit{Line Art}           & 0.93& \textbf{0.62}& 0.45& 0.93& 0.55& 0.38& 0.92& 0.56                  & 0.39& 0.89& 0.41                  & 0.26             & 0.90 & 0.46 & 0.30\\
\textit{Photo Depiction}    & 0.87& 0.41& 0.26& 0.89& \textbf{0.58}& 0.40& 0.89& 0.54                  & 0.37& 0.85& 0.45                  & 0.29             & 0.83 &  0.46 & 0.30\\
\textit{Art Depiction}      & 0.88& \textbf{0.49}& 0.33             & 0.86& 0.41& 0.26& 0.87& 0.43                  & 0.27& 0.87& 0.45                  & 0.29             & 0.84 & 0.36 & 0.22\\ \midrule
\textit{Stratified (all)}   & 0.91                  & 0.63            & 0.46             & 0.91                     & 0.60               & 0.43                & 0.91                     & 0.65               & 0.48                & 0.89                  & 0.59            & 0.42              & 0.87 & 0.53 & 0.36\\ \bottomrule
\end{tabular}%
}
\end{table*}

\section{Benchmark}
We first benchmark Segment Anything, as it is state-of-the-art for many segmentation tasks. We evaluate the model in six different prompting settings at the task of segmenting damage. 
Next, we evaluate several state-of-the-art semantic segmentation methods on our dataset, and compare them alongside linear probes of SAM and another state-of-the-art vision foundation model - DINOv2.
For rigorous evaluation in this supervised setting, we define a LOOCV protocol that measures how well models generalise to unseen media materials and contents, where not only the damaged, but the undamaged regions exhibit variance as a result of material and content properties.
We further evaluate two diffusion segmentation methods based on Stable Diffusion, demonstrating the dataset's versatility and underscoring the task's complexity. 
We analyze each method's effectiveness across Material and Content types within the dataset, finding that although some are capable of detecting damage within specific categories or through tailored prompting, no single method excels across the whole dataset, and all methods have different shortcomings in addressing our task.

\subsection{Zero-shot Segmentation with SAM}
\label{subsec:exp-sam}
\vspace{-0.52em}
In its intended zero-shot setting, SAM requires prompting which can be done in several ways (points, bounding boxes, masks, or text, though the latter is not publicly available), and which would not be available in a real-life damage restoration scenario. Furthermore, SAM is inherently not a semantic segmentation model; it is designed to segment individual visual entities without assigning semantic labels to them or grouping them, meaning that each damage instance must first be segmented individually and then manually assigned the correct label.

\subsubsection{Evaluation protocol}
\label{subsec:eval-protocol-sam}
We address these complexities and extensively benchmark SAM using an oracle to assign the correct label (Damaged or Clean) post-segmentation. We evaluate the model's ability to segment damage by deriving the required prompts from our ground-truth labels in several settings: point per segmentation label, point per segmentation instance (connected component), bounding box per segmentation label, and bounding box per segmentation instance. Additionally, we also evaluate prompting via a grid of points, which is the authors' original solution for cases where a prompt is unavailable, and prompting via text by employing Grounding SAM~\cite{ren2024grounded}, where the model is prompted with textual description of the damage types present in each image. We report macro-averaged metrics across the entire dataset as well as for each Material and Content split in Table~\ref{tab:sam}.

\subsection{Supervised Semantic Segmentation}
\label{subsec:exp-supervised}

We benchmark SOTA supervised semantic segmentation methods alongside linear probes trained over the features of vision foundation models on two tasks which require predicting pixel-wise segmentation maps: either for detecting damage (binary) or for detecting and classifying it (multiclass). 

\subsubsection{Evaluation protocol}\label{subsec:eval-protocol}

We split the data by Material and Content categories, performing a leave-one-out evaluation with one category left out for testing, and the rest split 8:2 into training and validation sets. This results in 14 splits: 10 for Material and 4 for Content. This approach assesses model performance on unseen media properties. For comparison, we train each model using a stratified split, ensuring all categories are in training, validation, and test sets. In total, each model is trained and evaluated 15 times across these settings.

\begin{table*}[t]
\tiny
\caption{Results across both Material and Content splits, for all baselines, trained on the task of multiclass damage segmentation. Best F1 scores for each test class in bold. Last row is a stratified  split over Material x Content.}
\label{tab:baselines-multiclass-semantic-segmentation}
\vspace{-8pt}
\resizebox{\linewidth}{!}{%
\begin{tabular}{@{}lccccccccccccccc@{}lll}
\toprule
\multirow{2}{*}{Test Class} & \multicolumn{3}{c}{\textbf{Segformer}}                     & \multicolumn{3}{c}{\textbf{\makecell{UPerNet + Swin}}}     & \multicolumn{3}{c}{\textbf{\makecell{UPerNet + ConvNeXt}}}            & \multicolumn{3}{c}{\textbf{\makecell{DINOv2 + MLP}}} &                   \multicolumn{3}{c}{\textbf{\makecell{SAM ViT-H + MLP}}}\\ 
\cmidrule(l){2-4} \cmidrule(l){5-7} \cmidrule(l){8-10} \cmidrule(l){11-13} \cmidrule(l){14-16} 
                            & \textbf{Acc} & \textbf{F1} & \textbf{mIoU} & \textbf{Acc} & \textbf{F1} & \textbf{mIoU} & \textbf{Acc} & \textbf{F1} & \textbf{mIoU} & \textbf{Acc} & \textbf{F1} & \textbf{mIoU}  & \textbf{Acc} & \textbf{F1} &\textbf{mIoU}  \\ \midrule
\textit{Wood}               & 0.14                  & 0.11& 0.08             & 0.14& 0.10& 0.08& 0.14& 0.11& 0.08& 0.23                  & \textbf{0.20}    & 0.15              & 0.05 & 0.06 & 0.05 \\
\textit{Ceramic}            & 0.18                  & 0.13& 0.10              & 0.17& 0.13& 0.10& 0.17& 0.12& 0.09& 0.28                  & \textbf{0.20}& 0.16              & 0.10 & 0.10 & 0.08\\
\textit{Textile}            & 0.10                   & 0.10             & 0.08             & 0.10                   & 0.10& 0.09& 0.09                  & 0.09& 0.08& 0.20                   & \textbf{0.18}   & 0.15              & 0.07 & 0.07 & 0.06 \\
\textit{Lime Plaster}       & 0.13                  & 0.12& 0.09             & 0.14                  & 0.14& 0.10& 0.18& 0.17& 0.12& 0.38                  & \textbf{0.30}    & 0.22              & 0.07 & 0.08 & 0.06\\
\textit{Canvas}             & 0.17                  & \textbf{0.17}   & 0.13             & 0.17& 0.16& 0.13& 0.18& \textbf{0.17}& 0.13& 0.18                  & \textbf{0.17}   & 0.13              & 0.06 & 0.07 & 0.05 \\
\textit{Tesserae}           & 0.07                  & 0.08& 0.06             & 0.09                  & 0.08& 0.07& 0.08& 0.09& 0.07& 0.16                  & \textbf{0.15}   & 0.12              & 0.06 & 0.06 & 0.05\\
\textit{Paper}              & 0.14                  & 0.11            & 0.09             & 0.17& \textbf{0.13}& 0.10& 0.16& 0.12& 0.10& 0.18                  & 0.11& 0.09              & 0.07 & 0.06 & 0.05 \\
\textit{Glass}              & 0.15                  & 0.15& 0.12             & 0.18& \textbf{0.19}& 0.14& 0.16& 0.17& 0.13& 0.15                  & 0.14& 0.11              & 0.07 & 0.07 & 0.05 \\
\textit{Film emulsion}      & 0.12                  & 0.07& 0.06             & 0.10& 0.08& 0.07& 0.11& \textbf{0.09}& 0.07& 0.13                  & 0.06            & 0.05              & 0.05 & 0.06 & 0.05\\
\textit{Parchment}          & 0.14                  & \textbf{0.14}   & 0.10              & 0.12                  & 0.12& 0.09& 0.15& 0.13& 0.10& 0.18                  & \textbf{0.14}   & 0.11              & 0.07 & 0.07 & 0.06\\
\midrule
\textit{Geom Patterns}      & 0.15                  & 0.08& 0.05             & 0.20& 0.17& 0.13& 0.17& 0.16& 0.12& 0.23                  & \textbf{0.19}   & 0.15              & 0.09 & 0.09 & 0.07 \\
\textit{Line Art}           & 0.15                  & \textbf{0.15}   & 0.11             & 0.18& 0.14& 0.12& 0.16& 0.13& 0.10& 0.20                   & 0.13            & 0.11              & 0.09 & 0.06 & 0.05\\
\textit{Photo Depiction}    & 0.13                  & 0.12& 0.09             & 0.13& 0.12& 0.10& 0.16& \textbf{0.13}& 0.10& 0.14                  & 0.10             & 0.08              & 0.07 & 0.06 & 0.04\\
\textit{Art Depiction}      & 0.12                  & 0.11& 0.09             & 0.13& 0.11& 0.09& 0.15& \textbf{0.14}& 0.10& 0.17                  & 0.12& 0.09 & 0.08 & 0.07 & 0.05 \\ \midrule
\textit{Stratified (all)}   & 0.21                  & 0.20             & 0.15             & 0.23                     & 0.26               & 0.21                & 0.17                     & 0.18               & 0.14                & 0.30                   & 0.25            & 0.18              & 0.12 & 0.12 & 0.09\\ \bottomrule
\end{tabular}%
}
\end{table*}

\subsubsection{Evaluated methods}

Due to the range of scales exhibited by different damage types with respect to the image plane, we chose to evaluate segmentation models which have been shown to excel at recognition at various scales:
\begin{itemize}[itemsep=0pt, topsep=0.3pt, parsep=0.2pt]
    \item \textbf{UPerNet}~\cite{xiao2018unified} is a framework which can leverage different vision backbones to learn from heterogeneous segmentation annotations. The flexibility of this framework allows us to evaluate both a convolutional variant, using a \textbf{ConvNeXt}\cite{liu2022convnet} backbone, and a transformer variant, using \textbf{Swin Transformer}~\cite{liu2021swin}.
    \item \textbf{SegFormer}~\cite{xie2021segformer} utilises a hierarchically structured Transformer encoder which outputs multiscale features, and combines those with a lightweight Multilayer Perceptron (MLP). 
    \item \textbf{DINOv2}~\cite{oquab2023dinov2} is a foundation vision model trained in a self-supervised setting, without any labels. The resulting features have been demonstrated to have emergent semantically meaningful properties. The authors utilise this by attaching a linear classifier, which is trained for semantic segmentation as a downstream task, achieving competitive results. We adopt this setup for our evaluation as well.
    \item \textbf{SAM}~\cite{kirillov2023segment} is a foundation vision model that can perform zero-shot image segmentation given prompts. To assess how useful its feature space is for the task of damage detection and segmentation, we train a linear probe over the largest available SAM vision encoder (ViT-H) in the same setting as for DINOv2.
\end{itemize}

For all except SAM and DINOv2, we fine-tune from ADE20K~\cite{zhou2017scene} weights.
Due to data imbalance towards Clean, we use macro-averaged Dice loss. We measure macro-averaged F1 Score and Mean Intersection over Union (mIoU), fine-tuning models until the validation F1 score has not improved for 10 epochs. For completeness, we also report macro-averaged Accuracy. Models are optimized with Adam~\cite{kingma2014adam}, using the best learning rates and weight decay from the respective papers.

\subsubsection{Data Augmentation}\label{subsec:data-aug}
Our dataset consists of images of varying high resolutions and aspect ratios. During training, we apply standard augmentations: random scaling, cropping to the model's input size, and flipping. For validation and testing, we split our images into $512\times512$ pixel overlapping patches, make predictions on each patch, and stitch them back together using Hann windows~\cite{pielawski2020introducing} before calculating the metrics.

\subsection{Diffusion-based segmentation}
\label{subsec:exp-diffusion-based}

In addition to the supervised methods, we also benchmark two recent diffusion-based segmentation approaches. Following DiffEdit~\cite{couairon2022diffedit}, we use contrasting predictions from a diffusion model conditioned on pairs of different single-word text prompts. Our positive prompts are "Flawless", "Unblemished", "Undamaged", paired with corresponding negative prompts "Damaged", "Deteriorated", "Defaced". Each image is evaluated against all nine resulting prompt pairs; we report  metrics for the pair with the best F1 score. This approach requires no training, relying on the learned prior of Stable Diffusion~\cite{rombach2021highresolution}. For our second approach, we benchmark DiffSeg~\cite{tian2024diffuse}, which is an unsupervised segmentation model relying on self-attention features extracted from Stable Diffusion; we set the KL-divergence threshold to $0.5$ in order to produce segments of higher granularity, and use an oracle to assign each predicted segment to the Damaged or Clean class, same as with SAM in~\ref{subsec:eval-protocol-sam}. Both methods are evaluated on inputs center-cropped and resized to $512\times512$; for DiffEdit we also provide results in full resolution in the Supplementary.

\begin{figure*}[t]
\begin{center}
    \includegraphics{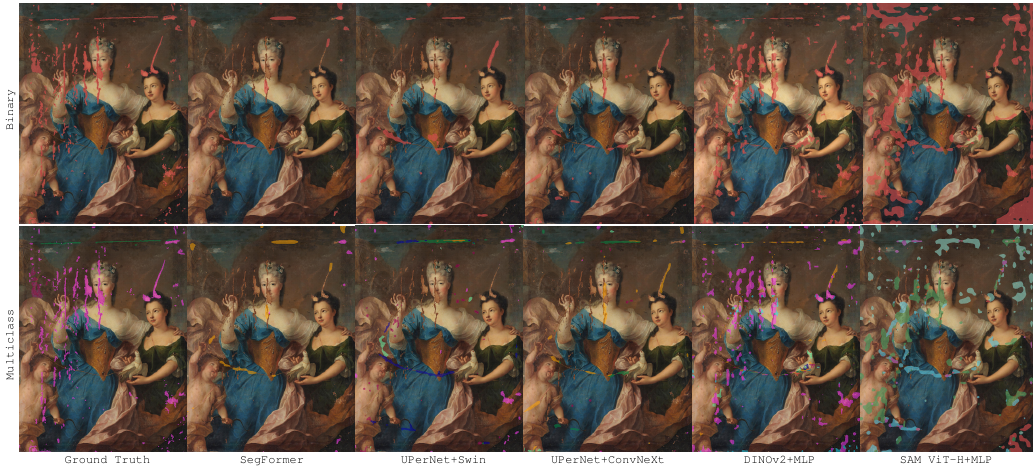}
\end{center}
\vspace{-10pt}
\caption{Qualitative comparison for supervised binary (top row) and multiclass (bottom row) damage segmentation.}
\label{fig:qualitative}
\end{figure*}

\subsection{Results}

Evaluation results are provided for all modes of prompting SAM in a zero-shot setting, all methods for the supervised semantic segmentation task (binary and multiclass setting) as well as for diffusion-based segmentation (text-guided and unsupervised). None of the benchmarked approaches manage to consistently perform well across different types of materials and contents. 

\paragraph{SAM is missing the point.}
We found that SAM can produce good segmentation results only \emph{if} prompted correctly (Figure~\ref{fig:sam-zero-shot}), but doing so requires an impractical number of prompts. As shown in Table~\ref{tab:sam}, using points-whether one per label or per artefact instance-yields poor results. Bounding boxes per class also perform poorly, especially when artefacts are spread across the image, as the boxes are too imprecise. Text-based prompts perform even worse, indicating that damage is not easily captured semantically. Point grid sampling also fails, as it doesn't guarantee all artefacts, especially fine ones, are sampled. The only method that provides adequate segmentation - achieving the highest average F1 score of $0.63$ - is using a bounding box for each artefact instance, but this requires an average of 112 boxes per image (up to 3407 for some), making it impractical for real-world use in terms of both human effort and computational resources, and essentially solving the detection task for SAM a priori. Furthermore, an oracle was needed to assign labels to the segmented regions.

\begin{table}[t]
\tiny
\centering
\caption{Results across both Material and Content splits diffusion based text-guided (DiffEdit) and unsupervised (DiffSeg) binary damage segmentation. 
}
\label{tab:diffusion}
\vspace{-8pt}
\resizebox{\linewidth}{!}{%
\begin{tabular}{@{}lcccccc@{}}
\toprule
\multirow{2}{*}{Test class} & \multicolumn{3}{c}{\textbf{\makecell{DiffEdit}}} & \multicolumn{3}{c}{\textbf{\makecell{DiffSeg}}} \\ \cmidrule(lr){2-4} \cmidrule(l){5-7} 
                            & \textbf{Acc}  & \textbf{F1}  & \textbf{mIoU}  & \textbf{Acc}  & \textbf{F1}  & \textbf{mIoU}  \\ \midrule
\textit{Wood}               & 0.77          & 0.13         & 0.07          & 0.90& 0.18& 0.15\\
\textit{Ceramic}            & 0.70           & 0.13         & 0.07          & 0.90& 0.39& 0.31\\
\textit{Textile}            & 0.79          & 0.19         & 0.11          & 0.93& 0.57& 0.47\\
\textit{Lime Plaster}       & 0.72          & 0.18         & 0.10           & 0.90& 0.42& 0.35\\
\textit{Canvas}             & 0.79          & 0.21         & 0.12          & 0.91& 0.28& 0.21\\
\textit{Tesserae}           & 0.76          & 0.11         & 0.06          & 0.92& 0.53& 0.45\\
\textit{Paper}              & 0.81          & 0.16         & 0.09          & 0.94& 0.29& 0.23\\
\textit{Glass}              & 0.78          & 0.19         & 0.11          & 0.94& 0.38& 0.30\\
\textit{Film emulsion}      & 0.78          & 0.08         & 0.04          & 0.98& 0.41& 0.43\\
\textit{Parchment}          & 0.83          & 0.13         & 0.07          & 0.93& 0.18& 0.15\\
\textit{Geom Patterns}      & 0.80           & 0.20          & 0.12          & 0.94& 0.45& 0.39\\
\textit{Line Art}           & 0.84& 0.14         & 0.08          & 0.95& 0.16& 0.13\\
\textit{Photo Depiction}    & 0.79          & 0.18         & 0.10           & 0.95& 0.36& 0.30\\
\textit{Art Depiction}      & 0.79          & 0.13         & 0.07          & 0.93& 0.30& 0.24\\ \bottomrule
\end{tabular}%
}
\vspace{-1.5em}
\end{table}

\paragraph{Supervised semantic segmentation.}
In the supervised setting, models can distinguish some damaged areas in the binary task, as seen in Figure~\ref{fig:qualitative} top row, but struggle with segmenting and classifying damage types in the multiclass setting for unseen contents or materials, regardless of pre-training or architecture, visualised in Figure~\ref{fig:qualitative} bottom row.
Table~\ref{tab:baselines-binary-semantic-segmentation} summarizes the binary segmentation results, where no model achieves an F1 score over $0.7$. Notably, DINOv2 outperforms SAM despite being a smaller model. In the multiclass setting (Table~\ref{tab:baselines-multiclass-semantic-segmentation}), all models perform poorly, with no F1 score exceeding $0.3$. Damage missclassifications are detailed in Section 3.1. of the Supplementary. DINOv2 performs best, likely due to its semantically meaningful features~\cite{oquab2023dinov2}, aiding in differentiating damage types. SAM, however, struggles to distinguish between damage types and produces imprecise segmentations when trained to segment all artefacts simultaneously, as it is unable to semantically group artefacts of the same type together within its feature space. This underscores the fact that the SAM is wholly inappropriate for this task, where multiple instances of damage may be present in the same image, and the model needs to meaningfully distinguish them.

\paragraph{Diffusion-based segmentation.}
In diffusion-segmentation approaches, we found that while models may localize or cluster damaged areas—such as in DiffSeg—the predictions are too imprecise for acceptable segmentation quality. As shown in Table~\ref{tab:diffusion}, the results are comparable to those from supervised models, with DiffEdit performing worse than DiffSeg. Qualitatively, DiffEdit sometimes identifies damaged areas but often highlights unrelated features, such as faces and high-frequency details, due to diffusion models' tendency to generate fine details at later steps~\cite{luo2024diffusion}. DiffSeg performs slightly better, aided by an oracle for labeling, but while it groups damage instances effectively due to the emergent semantic meaning found in Stable Diffusion's features - its segmentation remains imprecise since it relies on low-dimensional features.

\section{Conclusion}
Damage is ubiquitous in analogue media, particularly in heritage artifacts. While inpainting for restoration is relatively well-solved, reliably identifying damaged areas from digital images remains a significant challenge. We have introduced a damage taxonomy spanning diverse materials and contents to guide detection systems, and our extensive, permissively-licensed dataset sets a rigorous baseline for future work. Our benchmark results show that no state-of-the-art method performs acceptably in detecting damage. Supervised models fail to generalize to unseen materials and contents, while foundation models require excessive prompt engineering and still struggle to label damage accurately. There is substantial room to develop machine learning pipelines that can detect damage at a human-equivalent level, particularly for the pixel-perfect precision required in conservation. Our experiments highlight the complexity of the task, and we hope our dataset will inspire advancements in this area.

\section*{Acknowledgements}
This work was supported by the Engineering and Physical Sciences Research Council [grant number EP/R513222/1].

{\small
\bibliographystyle{ieee_fullname}
\bibliography{PaperForReview}
}

\end{document}